\pdfoutput=1

\documentclass[11pt]{article}

\usepackage[]{acl}

\usepackage{times}
\usepackage{latexsym}
\usepackage{xcolor}
\usepackage{stackengine}
\usepackage{amsmath,amssymb}
\newlength\lunderset
\newlength\rulethick
\usepackage{soul}
\usepackage{tikz}

\newcommand*\circled[1]{\tikz[baseline=(char.base)]{
            \node[shape=circle,draw,inner sep=2pt] (char) {#1};}}
\usepackage[T1]{fontenc}

\usepackage[utf8]{inputenc}

\usepackage{microtype}
\usepackage{graphicx}
\usepackage{amsmath}
\usepackage[linesnumbered, ruled, lined]{algorithm2e}
\usepackage{booktabs}
\usepackage{multirow}
\usepackage[normalem]{ulem}
\usepackage[symbol]{footmisc}

\usepackage{tabularx}
\newcounter{gaocomm} 
  
\definecolor{blue-violet}{rgb}{0.54, 0.17, 0.89}
\definecolor{mygreen}{rgb}{0.0, 0.5, 0.0}
\definecolor{awesome}{rgb}{1.0, 0.13, 0.32}
\definecolor{bostonuniversityred}{rgb}{0.8, 0.0, 0.0}

\title{OTExtSum: Extractive Text Summarisation with Optimal Transport}

\author{Peggy Tang$^{\dagger}$, Kun Hu$^{\dagger}$, Rui Yan$^{\star}$, Lei Zhang$^{\diamond}$, Junbin Gao$^{\ddagger}$, Zhiyong Wang$^{\dagger}$ \\[-0.1cm]

$^{\dagger}$School of Computer Science, The University of Sydney   \\[-0.2cm] $^{\star}$Gaoling School of Artificial Intelligence,  Renmin University of China \\[-0.2cm] $^{\diamond}$International Digital Economy Academy \\[-0.2cm] $^{\ddagger}$The University of Sydney Business School, The University of Sydney \\[-0.2cm]

\tt \{peggy.tang,junbin.gao,zhiyong.wang\}@sydney.edu.au, \\[-0.3cm] \tt kuhu6123@uni.sydney.edu.au,\\[-0.3cm] \tt ruiyan@ruc.edu.cn,leizhang@idea.edu.cn
}

\begin{document}
\maketitle
\begin{abstract}
Extractive text summarisation aims to select salient sentences from a document to form a short yet informative summary. 
While learning-based methods have achieved promising results, they have several limitations, such as dependence on expensive training and lack of interpretability. 
Therefore, in this paper, we propose a novel non-learning-based method by for the first time formulating text summarisation as an Optimal Transport (OT) problem, namely Optimal Transport Extractive Summariser (OTExtSum). 
Optimal sentence extraction is conceptualised as obtaining an optimal summary that minimises the transportation cost to a given document regarding their semantic distributions. 
Such a cost is defined by the Wasserstein distance and used to measure the summary's semantic coverage of the original document. 
Comprehensive experiments on four challenging and widely used datasets -  MultiNews, PubMed,  BillSum, and CNN/DM  demonstrate that our proposed method outperforms the state-of-the-art non-learning-based methods and several recent learning-based methods in terms of the ROUGE metric. 
\footnote[1]{Our code is publicly available for research purpose in  https://github.com/peggypytang/OTExtSum/}

\end{abstract}

\begin{figure}[!h]
\centering
\includegraphics[width=.40\textwidth]{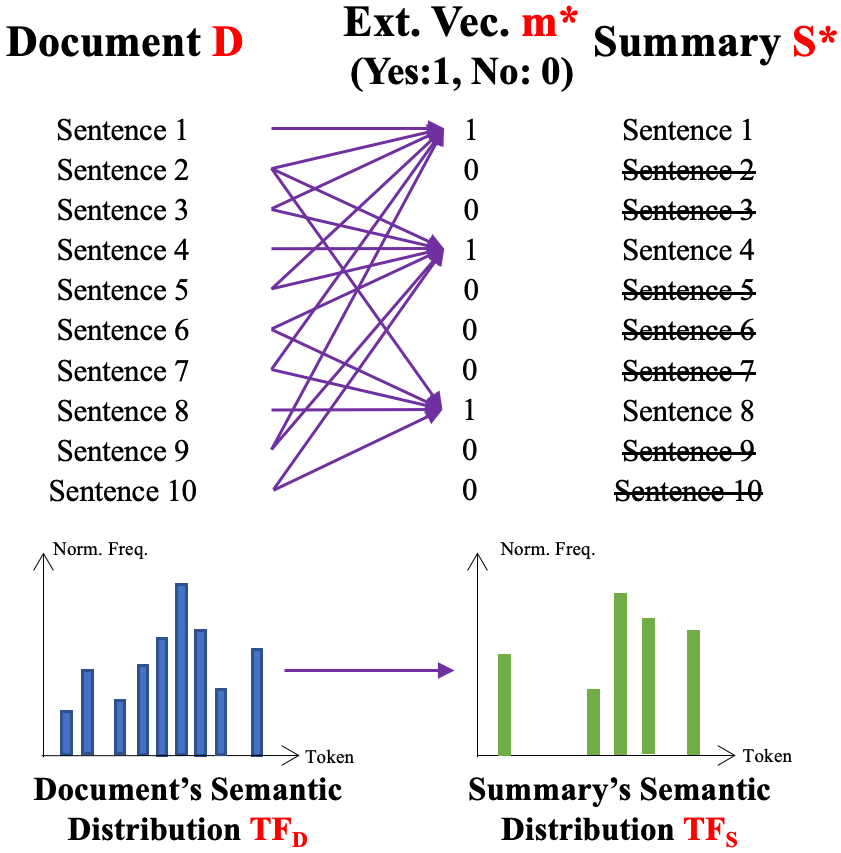} 
\caption{Illustration of Optimal Transport Extractive Summariser (OTExtSum): the formulation of extractive summarisation as an optimal transport (OT) problem. Optimal sentence extraction is conceptualised as obtaining the optimal extraction vector $\mathbf{m^*}$, which achieves an OT plan from a document $\mathbf{D}$ to its optimal  summary $\mathbf{S^*}$ that has the minimum transportation cost. Such a cost is defined as the Wasserstein distance between the document's semantic distribution $\text{TF}_\mathbf{D}$ and the summary's semantic distribution $\text{TF}_\mathbf{S}$ and is used to measure the summary's semantic coverage.
}
\label{fig:otextsum}
\end{figure}



\section{Introduction}

Text summarisation aims to condense a given document into a short and succinct summary that best covers the semantics of the document with the least redundancy. 
It helps users quickly browse and understand long documents by focusing on their most important sections \cite{IMani2001automatic,ANenkova2011automatic}. A common practice for text summarisation is extractive summarisation which aims to select the salient sentences of a given document to form its summary. 
Extractive summarisation ensures the production of grammatically and factually correct summaries, though the output summaries could be inflexible. Since abstractive summaries are highly prone to contain contents that are unfaithful and nonfactual to the original document \cite{JMaynez2020Faithfulness}, extractive summaries are more practical for real-world scenarios, especially for the domains requiring formal writing such as legal, science, and journalism documents. 

Existing methods \cite{yao2017recent} often first score the importance of individual sentences of a given document and then combine the top-ranked ones to form a summary. However, the sentences with high importance scores may not well represent the document from a global perspective, which results in a sub-optimal summary. 
Recently, learning-based methods, especially those based on 
supervised and unsupervised deep learning techniques \cite{SNarayan2018Ranking,HZheng2019PacSum, XZhang2019HIBERT,SNarayan2020Stepwise,JXu2020Discourse,zhong2020extractive, VPadmakumar2021unsupervised} can significantly improve summarisation performance. However, training deep learning models is computationally expensive, and it can be difficult to apply those models learned from a particular domain to other domains with different distributions. Moreover, deep learning methods generally lack interpretability for the summarisation process.

Motivated by these issues, we propose a novel non-learning based extractive summarisation method, namely Optimal Transport Extractive Summariser (OTExtSum). As illustrated in Figure \ref{fig:otextsum}, we formulate extractive summarisation based on the optimal transport (OT) theory \cite{peyre2019computational}. 
A candidate summary can be evaluated by an OT plan regarding the optimal cost to transport between the semantic distributions of the summary and its original document. Then a Wasserstein distance can be obtained with this optimal plan to measure the discrepancy between the two distributions. 
To this end, it can be expected that a summary of high quality minimizes this Wasserstein distance. 
Moreover, a common assumption in the formulations of the OT problem  is that the source and target distributions are fixed. 
In OTExtSum problem formulation, we relax this assumption by adding an extraction vector $\mathbf{m^*}$ to indicate which document sentences would be extracted to form the summary's semantic distribution, thus making the target distribution variable.

The semantic distributions of a given document and its candidate summary can be formulated in line with the frequency of their tokens. Inspired by Word Mover's Distance \cite{MKusner2015WMD}, summarisation can be conceptualized as moving the "semantics" of a given document to its summary, and the ideal summary is obtained at the minimal transportation cost. This ensures the highest semantic coverage of the given document and the least redundancy in the summary without explicitly modelling conventional criteria such as relevance and redundancy.
Thus, under the OT plan, 
the Wasserstein distance indicates the candidate summary's semantic coverage of the given document. 

We design two optimisation strategies to approximate the extraction vector $\mathbf{m^*}$, 
namely beam search strategy \cite{CTillmann2003Word}, which iteratively evaluates the semantic coverage scores of a set of candidate summaries to obtain the optimal extraction, and binary integer programming strategy, which approximates the optimal extraction given the constraints of the Wasserstein distance and extraction budget. As a non-learning based method, OTExtSum does not require any training and is applicable to different document domains. Furthermore, it provides explainable results in terms of the semantic coverage of the summary. 


There have been some studies on OT in NLP, such as document distance \cite{MKusner2015WMD, yurochkin2019hierarchical}, text generation \cite{chen2018adversarial}, text matching \cite{swanson2020rationalizing}, and machine translation \cite{Xu2021Vocabulary}. These methods generally focus on deriving similarities between words, sentences, and documents. On the contrary, we for the first time formulate text summarisation as an OT problem that optimally transports the semantic distributions between two texts (e.g., source document and summary candidate).

Overall, the key contributions of this paper are:
\begin{itemize}
	\item We propose a non-learning based extractive summarisation method - OTExtSum by treating the text summarisation task as an optimal transport problem for the first time.  
	\item We design two optimisation strategies for OTExtSum
	: beam search strategy and binary integer programming strategy.
	\item We present an interpretable visualisation of the semantic coverage of a generated summary by visualising the transport plan between summary tokens and document tokens.
	\item Comprehensive experimental results on four widely used datasets, including CNN/DM, MultiNews, BillSum and PubMed, demonstrate that OTExtSum outperforms the state-of-the-art non-learning based methods. 
\end{itemize}


\section{Related Work}
\label{litreview}

Generally, text summarisation methods can be categorized as extractive, abstractive, and hybrid ones. 
While abstractive and hybrid summarisation methods \cite{LLebanoff2019Scoring,JZhang2020pegasus} aim to mimic human beings in summarisation by paraphrasing a given document, extractive summarisation generally produces more factual summaries. 
In this section, we review existing extractive summarisation methods in two categories: non-learning based and learning-based methods. 

\subsection{Non-learning based Methods}

Most of the non-learning based methods conceptualise text summarisation as a sentence ranking task. Each sentence in a given document is scored in terms of various sentence importance criteria, which measure how well the sentence could represent the document. The top-ranked sentences are combined to form a summary. 
These methods often heavily rely on handcrafted features in regards to linguistic knowledge by focusing on local and/or global contexts. 

\textbf{Local Context based Methods.} 
Local context-based methods rank a sentence based on the features obtained from the sentence itself. Sentence features such as frequency-based and topic-based were studied. Frequency-based features \cite{HEdmundson1969New, EHovy1998Summarist} assume that the occurrence of high-frequency terms in a sentence is associated with their importance.  
Topic-based features \cite{JKupiec1995trainable, CNobata2004crl, CLin2000Topic} assume that the density of a set of topic terms is highly correlated to the topic of a document. 

\textbf{Global Context based Methods.} 
As local context features could overlook the correlations between sentences and lead to redundant summaries involving similar sentences, global context-based methods rank individual sentences from the perspective of the entire document. 
Discourse-based methods \cite{DMarcu1999discoursetrees} construct a document's rhetorical structure and extract the sentences on the longest chain of the semantic structure, i.e. the main topic.  Centroid-based methods \cite{DRadev2000Centroid} cluster the sentences of a document through similarity measures and rank the sentences based on their distances to the cluster centroids. TextRank \cite{RMihalcea2004Textrank}, as a graph-based method,  is the state-of-the-art non-learning based method. 
A graph among document sentences
is first formed by connecting sentences using sentence similarity scores, then the sentence connectivity can be used to score the importance of a sentence. Nonetheless, the nature of these sentence based scoring methods could miss summary-level or document-level patterns. 

\subsection{Learning-based Methods}

Instead of utilising handcrafted features, due to the great success of deep learning in many natural language processing tasks, recent studies on extractive summarisation aim to learn sentence features from the corpus in a data-driven manner.  

\textbf{Supervised Methods.} Most of these methods follow the sentence ranking conceptualisation, and an encoder-decoder scheme is generally adopted
\cite{RNallapati2017Summarunner,XZhang2019HIBERT,SNarayan2020Stepwise,JXu2020Discourse}. An encoder formulates document or sentence representations, and a decoder predicts a sequence of sentence importance scores with the supervision of ground-truth sentence labels. 

\textbf{Reinforcement Learning based Methods.} 
Reinforcement learning (RL) can be utilised for extractive summarisation 
by directly optimising the ROUGE metric, which is used as the training reward. The RL based summarisation task can be treated as a sentence ranking problem similar to the aforementioned methods \cite{SNarayan2018Ranking} or as a contextual-bandit problem \cite{
LLuo2019Reading} . 

\textbf{Unsupervised Methods} Various unsupervised methods have also been proposed to leverage pre-trained language models to compute sentence similarities and select important sentences. Some methods \cite{HZheng2019PacSum} use these similarities to construct a sentence graph and select sentences based on their centrality. Some methods \cite{VPadmakumar2021unsupervised} use these to score relevance and redundancy of sentences as selection criteria. 

Although these learning-based methods have significantly improved summarisation performance, 
computationally expensive training costs are inevitable, and it is challenging to generalise the trained models to documents from other domains that have distributions different from the training dataset. 
In addition, it is difficult to explain the correspondence and the coverage between a summary and a source document using these deep models. Therefore, to address these limitations, we revisit the non-learning based approach and propose a novel summarisation method by exploring the optimal transport theory for the first time. 

\section{Methodology}
\label{methods}


As shown in Figure \ref{fig:otextsum}, OTExtSum utilizes a text OT approximation to obtain  the  optimal  extraction  vector $\mathbf{m^*} = [m_1,...,m_n]^T$, where $m_i \in \{0,1\}$ denotes whether the $i$-th sentence is to be extracted (denoted by 1) or not (denoted by 0). The optimal extraction vector $\mathbf{m^*}$ achieves an OT plan from the semantic distribution of the document to that of its optimal candidate summary which has the minimum total transportation cost.

The OT approximation consists of four components: 
1) a tokeniser \& embedding procedure that formulates token level representations and a semantic distribution estimation that computes the frequency of each token within a summary or a document ; 2) a transportation cost matrix that measures the cost using one token to represent another based on their Euclidean distances; 3) an OT solver that approximates Wasserstein distance and semantic coverage of the candidate summaries; and 4) 
an optimisation strategy that obtains the optimal extraction vector by choosing the summary with the minimum Wasserstein distance, and thus with the highest semantic coverage of the source document.


\subsection{Optimal Transport} \label{sec:ot}

Consider a transportation problem that transports goods from a collection of suppliers \(\mathbf{D}=\{d_{i}|i=1,...,N\}\) to a collection of customers \(\mathbf{S}=\{s_{j}|j=1,...,N\}\), where \(d_{i}\) and \(s_{j}\) indicate the supply quantity of the \(i\)-th supplier and the order quantity of the \(j\)-th customer, respectively. 
Note that, in this study, we consider the number of suppliers to be the same as the customers. 
By defining $t_{ij}$ as the quantity transported from the $i$-th supplier to the $j$-th customer, a transport plan $\mathbf{T}=\{t_{ij}\}\in\mathbf{R}^{N\times N}$ can be obtained. 
Given a cost matrix $\mathbf{C}=\{c_{ij}\}\in\mathbf{R}^{N\times N}$, where $c_{ij}$ is the cost to deliver a unit of goods from the $i$-th supplier to the $j$-th supplier, the cost of the transport plan $\mathbf{T}$ can be calculated. 
Particularly, an OT plan $\mathbf{T}^{*}=\{t^{*}_{i,j}\}\in\mathbf{R}^{N\times N}$ in pursuit of minimising the transportation cost can be obtained by solving the following optimisation problem: 
\begin{equation} \label{eqt:OT}
\begin{aligned}
&\mathbf{T}^{*} = \underset{\mathbf{T}}{\text{argmin}} \sum_{i,j=1}^{N} t_{ij}c_{ij}, \\
\text{s.t.} \; &\sum_{j=1}^{N}t_{ij}=d_{i},\;\;\; \forall i \; \in \; \left \{1,...,N \right \}, \\ &\sum_{i=1}^{N}t_{ij}=s_{j},\;\;\; \forall j \; \in \; \left \{1,...,N  \right \}, \\
&t_{ij}\geq 0,\;\;\; \forall i, j \; \in \; \left \{1,...,N \right \},
\end{aligned}
\end{equation} 
where the first two constraints indicate the quantity requirements for both suppliers and customers and the last constraint proves a non-negative order quantity. Mathematically, this OT problem is to find a joint distribution $\mathbf{T}$ with respect to a cost $\mathbf{C}$, of which the marginal distribution is $\mathbf{D}$ and $\mathbf{S}$. In particular, Wasserstein distance can be defined as: 
\begin{equation} \label{eqt:WD}
d_W(\mathbf{D},\mathbf{S}|\mathbf{C}) = \sum_{i,j} t^{*}_{i,j}c_{i,j}.
\end{equation}
It can be viewed as the distance between the two probability distributions $\mathbf{D}$ and $\mathbf{S}$, if they are normalized, in line with the cost $\mathbf{C}$.



\subsection{Semantic Distribution} \label{sec:procedure1}


In the context of text summarisation, denote $\mathbf{D}=\{\mathbf{s}_1, ... , \mathbf{s}_n\}$ to represent a document, where $\mathbf{s}_i$ denotes the $i$-th sentence contained in the document. The sentence $\mathbf{s}_i$ has a semantic distribution  $\text{TF}_i \in \mathbb{R}^N$ computed by the normalised bag-of-tokens with removal of stop-words:

\begin{equation}
\begin{aligned}
\label{eqt:TFsentence}
\text{TF}_i &= [TF_{i1}, ... , TF_{iN}]^T, 
\\ TF_{ij} &= \frac{d_{j}}{\sum_{k=1}^{N} d_{k}},
\end{aligned}
\end{equation} 
where $d_j$ indicates the count of the $j$-th token in a vocabulary of size $N$.

A document $\mathbf{D}$ has a semantic distribution $\text{TF}_\mathbf{D}$:
\begin{equation}
\begin{aligned}
\label{eqt:TFD}
\text{TF}_\mathbf{D} &= \frac{\text{TF}_1 + … + \text{TF}_n}{n}.
\end{aligned}
\end{equation} 

For a summary $\mathbf{S}\subset \mathbf{D}$ with its corresponding extraction vector $\mathbf{m}$, of which the $i$-th element $m_i$ is an indicator ($m_i=1$ if $\mathbf{s}_i\in\mathbf{S}$, $m_i=0$ otherwise), it has a semantic distribution $\text{TF}_{\mathbf{S}}$:

\begin{equation}
\begin{aligned}
\label{eqt:TFS}
\text{TF}_{\mathbf{S}} = \frac{m_1 \times \text{TF}_1 + ... + m_n \times \text{TF}_n   }{m_1+...+m_n}.
\end{aligned}
\end{equation} 


In our proposed method
, a normalization step is introduced to approximate the semantic distributions of $\mathbf{D}$ and $\mathbf{S}$ with term frequency. 
Note that after the normalization $\text{TF}_{\mathbf{D}}$ and $\text{TF}_{\mathbf{S}}$ have an equal total good quantities of \(1\) and can be \textit{completely transported from one to the other}. In addition, $\text{TF}_{\mathbf{D}}$ and $\text{TF}_{\mathbf{S}}$ satisfy the property of discrete probability distributions, of which the sum should be 1. 

\subsection{Transport Cost between Tokens} \label{sec:procedure2}

We define the unit transportation cost between two tokens by measuring their semantic similarity. Intuitively, the more semantically dissimilar a pair of tokens are, the higher the ``transport cost" of transporting one token to another. 
Given a pre-trained tokeniser and token embedding model with $N$ tokens, define $\mathbf{v}_i$ to represent the feature embedding of the $i$-th token. The transport cost from the $i$-th token to the $j$-th token $c_{ij}$ in $\mathbf{C}$ can be written as:
\begin{equation} \label{eqt:costfunction}
c_{ij} = \left \|\mathbf{v}_{i} - \mathbf{v}_{j}\right \| _{2} ,
\end{equation}
which is based on the Euclidean distance. \footnote{We investigated the effect of different distance measurements. As discussed in Section \ref{sec:quantitative}, cost matrix based on the Euclidean  distance and the cosine distance yield similar ROUGE scores.}


\subsection{Semantic Coverage of Candidate Summaries} \label{sec:procedure3}

Intuitively, a good summary $\mathbf{S}$ is supposed to be close to the document $\mathbf{D}$ in terms of their semantic distributions. 
OTExtSum utilizes the Wasserstein distance to measure the distance between the two associated semantic distributions $\text{TF}_{\mathbf{D}}$ and $\text{TF}_{\mathbf{S}}$ with the OT cost. The computation of the Wasserstein distance has time complexity of $O(p^3\text{log}(p))$ \cite{altschuler2017near}, where $p$ denotes the number of unique words in the document. 

In detail, it can be obtained with Eq. (\ref{eqt:WD}) 
as $d_W(\text{TF}_{\mathbf{D}},\text{TF}_{\mathbf{S}}|\mathbf{C})$ with a pre-defined cost matrix $\mathbf{C}$. Then a semantic coverage score of the summary $\mathbf{S}$ in respect to the document $\mathbf{D}$ can be further defined based on the Wasserstein distance: 
\begin{equation} \label{eqt:coverage}
g(\mathbf{D}, \mathbf{S}) = 1-d_W(\text{TF}_{\mathbf{D}},\text{TF}_{\mathbf{S}}|\mathbf{C}).
\end{equation}
Therefore, OTExtSum aims to search for an extraction vector $\mathbf{m}$, of which the corresponding summary $\mathbf{S}$ minimises the Wasserstein distance, i.e. maximising the semantic coverage score for the given document $\mathbf{D}$ by solving OT problems.

\subsection{Optimisation Strategy} \label{sec:procedure4}


The remaining problem for OTExtSum is to search for the optimal extraction vector $\mathbf{m^*}$ which achieves the minimum total transportation cost from the semantic distribution of the document $\text{TF}_{\mathbf{D}}$ to that of the optimal summary $\text{TF}_{\mathbf{S}}$, given a budget \(B\) which is the number of sentences can be extracted to create a summary:

\begin{equation} \label{eqt:optimisation}
\begin{aligned}
\mathbf{m}^{*} = \underset{\mathbf{m}}{\text{argmin }} d_W&(\text{TF}_{\mathbf{D}},\text{TF}_{\mathbf{S}}|\mathbf{C}),
\\\text{s.t.  } m_1 &+...+m_n \leq B.
\end{aligned}
\end{equation} 

In search of optimal extraction vector $\mathbf{m^*}$, we design two optimisation strategies, namely beam search strategy to achieve better coverage approximation, and binary integer programming strategy to achieve better computational efficiency.


\begin{algorithm}[h!]
\SetKwInOut{Input}{Input}\SetKwInOut{Output}{Output}
\Input{$\mathbf{D}$ the document, $B$ the budget of the number of extracted sentences, $K$ the beam width. }
\Output{$\mathbf{S}^{*}$ the optimal extractive summary. }
\BlankLine

{Compute the cost matrix $\mathbf{C}$, }
{and the document's semantic distribution $\text{TF}_{\mathbf{D}}$;}

{Initialise $\mathbf{m}=\mathbf{0}$, i.e. the candidate summary set $\mathbb{S}=\varnothing$;}

\While(\tcp*[h] Beam search){\# of sentences in candidate summary $\le B$ 
;}{ 
    
    \For{$k=1,...,|\mathbb{S}|$}{
    
        \emph{Generate the successor set $\mathbb{S}^{k}_{b}$ for $\mathbf{S}^k\in\mathbb{S}$;}
    }
    
   {$\mathbb{S} \leftarrow \bigcup_{k} \mathbb{S}^{k}_{b}$;}
    
    \For{$k=1,...,|\mathbb{S}|$}{
    
        \emph{Compute the semantic distribution $\text{TF}_{\mathbf{S}^{k}}$ of $\mathbf{S}^k\in\mathbb{S}$;}
        
        \emph{Compute the Wasserstein distance $d_W(\text{TF}_{\mathbf{D}},\text{TF}_{\mathbf{S}^k}|\mathbf{C})$ and the semantic coverage $g(\text{TF}_{\mathbf{D}},\text{TF}_{\mathbf{S}^k}|\mathbf{C}))$;}
        

    }
   {Keep the top $K$ candidate summaries with the highest $g(\text{TF}_{\mathbf{D}},\text{TF}_{\mathbf{S}^k}|\mathbf{C}))$\ and prune the rest in $\mathbb{S}$;}
    
}

{$\mathbf{S}^{*} = \underset{\mathbf{S}^{k}\in \mathbb{S}}{\text{argmax }} g(\text{TF}_{\mathbf{D}},\text{TF}_{\mathbf{S}^k}|\mathbf{C}))$;}


\caption{Optimisation of OTExtSum with Beam Search Strategy}\label{algo:OTExtSumBeam}
\end{algorithm}

\subsubsection{Beam Search Strategy}

The Beam Search (BS) strategy with the beam width \(K\) maintains the candidate summary set $\mathbb{S}$ and searches for the optimal extraction vector $\mathbf{m^*}$, thus the optimal extractive summary $\mathbf{S^*}$. Algorithm \ref{algo:OTExtSumBeam} presents the steps to obtain the optimal summary with OTExtSum using the BS strategy. The time complexity is $O(BKn(p^3\text{log}(p)))$. 

Initially, we have $\mathbf{m} = \mathbf{0}$, where none of the sentences are extracted. Then, each sentence in the document $\mathbf{D}$ is selected as a candidate summary, which derives a set of candidate extraction vectors corresponding to a set of candidate summaries, and its semantic coverage score can be evaluated. The top $K$ candidate summaries in terms of the semantic coverage are kept in the set $\mathbb{S}$ and the rest are pruned. 
During the $b$-th iteration of the beam search, by appending each possible sentence to an existing candidate summary $\mathbf{S}^{k} \in \mathbb{S}$, where the sentence is not in $\mathbf{S}^{k}$, a set of new candidate summaries $\mathbb{S}^{k}_{b}$ can be obtained. Then $\mathbb{S}$ is updated by combining all these sets of new candidate summaries in regards to $k$: 
\begin{equation}
    \mathbb{S} \leftarrow \bigcup_{k} \mathbb{S}^{k}_{b}.
\end{equation}
At the end of beam search, a set of final $K$ summary candidates within the budget \(B\) is obtained. 

Among the $K$ final candidates from the beam search, OTExtSum obtains the optimal extraction vector and thus the optimal summary by choosing the candidate with the highest semantic coverage of the document $\mathbf{D}$. 

\begin{algorithm}[h!]
\SetKwInOut{Input}{Input}\SetKwInOut{Output}{Output}
\Input{$\mathbf{D}$ the document, $B$ the budget of the number of extracted sentences, $T$ the number of iterations. }
\Output{$\mathbf{S}^{*}$ the optimal extractive summary.  }
\BlankLine

{Compute the cost matrix $\mathbf{C}$, }
{Compute document's semantic distribution $\text{TF}_{\mathbf{D}}$;}

{Initialise $\mathbf{w} \in \mathbb{R}^{n}$;}

\For{iteration $t \in [1,...,T]$}{
    \emph{Convert $\mathbf{w}$ to probability value $\mathbf{pr}$ with Sigmoid function;}
    
    \emph{Convert $\mathbf{pr}$ to $\mathbf{b} = [b_i,..,b_n]$ by hard sampling from the Gumbel-Softmax distribution;}
    
    \emph{Construct summary's semantic distribution $\text{TF}_{\mathbf{S}}$;}
    
    \emph{Compute the Wasserstein distance $d_W(\text{TF}_{\mathbf{D}},\text{TF}_{\mathbf{S}}|\mathbf{C})$;}
    
    \emph{Compute the $L_1$ regularisation of $\mathbf{b}$;} 
    
    \emph{Compute loss by weighted sum of the Wasserstein distance and the squared difference of $B$ and $\mathbf{b}$;}
    
    \emph{Compute gradients and update $\mathbf{w}$;}
}

{Compute $\mathbf{m^*}$ by soft sampling $\text{Sigmoid}(\mathbf{w})$ from the Gumbel-Softmax distribution; }

{Obtain $\mathbf{S}^{*}$ by extracting top-$B$ sentences with the highest $m_i$ values for $i = 1,...,n$;}

\caption{Optimisation of OTExtSum with Binary Integer Programming Strategy}\label{algo:OTExtSumBIP}
\end{algorithm}

\subsubsection{Binary Integer Programming Strategy}

Some prior works showed that integer linear programming is an efficient solution to summarisation problem \cite{RMcDonald2007globalinference, DGillick2009scalable}.
The Binary Integer Programming (BIP) strategy therefore is utilised to search for the optimal extraction vector $\mathbf{m^*}$ with \(T\) iterations. Based on the extraction vector, we obtain the optimal extractive summary $\mathbf{S^*}$. Algorithm \ref{algo:OTExtSumBIP} presents the optimisation steps to obtain the optimal summary with OTExtSum using the BIP strategy. The time complexity is $O(T(p^3\text{log}(p)))$.

As $\mathbf{m^*}$ is a multi-hot vector and is not differentiable, to make the backpropagation work, we optimise a proxy continuous vector $\mathbf{w} \in \mathbb{R}^{n}$, which is differentiable. Then we hard sample from the Gumbel-Softmax distribution \cite{maddison2016gumbel} to discretise and compute a multi-hot vector $\mathbf{b}$ during the iterations, and soft sample to compute $\mathbf{m^*}$ at the end.

The BIP strategy optimises the following loss function w.r.t. $\mathbf{w}$, which is a weighted sum of the Wasserstein distance $d_W(\text{TF}_{\mathbf{D}},\text{TF}_{\mathbf{S}})$ and the $L_1$ regularisation of $\mathbf{b}$ \footnote{We choose $L_1$ regularisation for sparsity \cite{ANg2004feature}. }:
\begin{equation}
d_W(\text{TF}_{\mathbf{D}},\text{TF}_{\mathbf{S}}|\mathbf{C}) + \alpha |B - \sum_{i=1}^{n} b_{i}|, 
\end{equation}
where $\alpha$ denotes the weight of $L_1$ regularisation.

\section{Experimental Results and Discussions}
\label{sec:setup}

\subsection{Datasets}

To validate the effectiveness of the proposed OTExtSum on the documents with various writing styles and its ability to achieve improved summarisation performance, we perform experiments on four widely used challenging datasets collected from different domains.

\begin{table}[!h]
  \resizebox{.48\textwidth}{!}{
    \begin{tabular}{|l||cccc|}
    \hline
    \textbf{Dataset}  & \textbf{Multi-News} & \textbf{BillSum} & \textbf{PubMed} & \textbf{CNN/DM}\\
    \hline\hline
    \textbf{Domain} & News  & Law   & Science  & News\\
    \textbf{\#Sent./Doc.}  & 80    & 46      & 102 & 33\\
    \textbf{$B$}  &  9   & 7     & 6 &  3\\
    \textbf{Test Set Size}  & 5,622   &   3,269   & 6,658 & 11,490\\
    \hline
    \end{tabular}%
    }
    \caption{Overview of the datasets. \#Sent./Doc. denotes the average number of sentences in the documents, $B$ denotes the budget of number of extracted sentences.}
  \label{tab:dataset}%
\end{table}%

\begin{table*}[h]
\small
\resizebox{.99\textwidth}{!}{
\begin{tabular}{|l||ccc|ccc|ccc|ccc|}
\hline
\multicolumn{1}{|l||}{\textbf{Method}} & \multicolumn{3}{c|}{\textbf{Multi-News}} & \multicolumn{3}{c|}{\textbf{BillSum}} & \multicolumn{3}{c|}{\textbf{PubMed}} & \multicolumn{3}{c|}{\textbf{CNN/DM}}\\
\multicolumn{1}{|l||}{}                                                              & ROUGE-1     & ROUGE-2     & ROUGE-L    & ROUGE-1    & ROUGE-2    & ROUGE-L    & ROUGE-1    & ROUGE-2   & ROUGE-L   & ROUGE-1    & ROUGE-2   & ROUGE-L \\ \hline\hline
LEAD                                            & 42.3       & 14.2       & 22.4      & 43.5       & 25.6      & 37.8      & 34.0      & 8.6      & 27.1  & 40.0      & 17.5     & 32.9    \\  \hline
ORACLE                                       & 45.4       & 20.6 & 28.1      & 43.7     & 25.7      & 38.0      & 37.1      & 15.5      & 30.4  & 43.1      & 23.7      & 37.5    \\ \hline  \hline
\multicolumn{13}{|l|}{\textbf{Non-learning based Methods}}  \\ \hline\hline
LSA   \cite{YGong2001LSA}                                         & -       & -       & -     & 32.6      & 15.7      & 26.3      & 33.9      & 9.9     & 29.7  & -      & -      & -  \\  \hline
LexRank     \cite{GErkan2004LexRank}                                       & 38.3       & 12.7       & 13.2      & -    & -      & -     & 39.2      & \textbf{13.9}      & \textbf{34.6}  & -     & -      & -   \\  \hline
TextRank \cite{RMihalcea2004Textrank}                                           & 38.4     & 13.1     & 13.5      & 34.4     & 17.8     & 27.8     & -      & -      & -  & \ul{34.1}     & \textbf{12.8}      & 22.5   \\  \hline
\textbf{OTExtSum-BIP (GPT2)}                                            & 40.6      & 12.1       & 20.7     & 36.6   & 15.6      & 30.6      & 35.4     & 10.8    & 28.8 & \ul{34.1}    & \ul{12.6}      & \textbf{28.1}  \\  \hline
\textbf{OTExtSum-BIP (BERT)}                                            & 40.6       & 12.1      & 20.7      & 36.6    & 15.6    & 30.6     & 35.4     & 10.8     & 28.8  & \ul{34.1}      & \ul{12.6}     & \textbf{28.1}   \\  \hline
\textbf{OTExtSum-BS (Word2Vec)}                                            & 42.3      & 12.8       & 21.9      & \textbf{40.1}      & \ul{19.4}     & \textbf{34.3}      & 38.2      & 11.7      & 30.8  & 32.3      & 10.8      & 25.9   \\  \hline
\textbf{OTExtSum-BS (GPT2)}                                            & \ul{42.4}       & \textbf{14.2}       & \textbf{23.2}      & 36.5      & \textbf{19.7}      & 32.0      & \ul{39.7}      & \ul{13.8}      & 32.3  & 33.5      & 12.0      & 26.7   \\  \hline
\textbf{OTExtSum-BS (BERT)}                                            & \textbf{43.1}       & \ul{13.9}       & \ul{22.5}      & \ul{37.5}      & \textbf{19.7}      & \ul{32.6}      & \textbf{39.8}      & 13.6      & 32.3  & \textbf{34.5}      & \textbf{12.8}      & \ul{27.8}   \\ 
\hline  \hline
\multicolumn{13}{|l|}{\textbf{Unsupervised Deep Learning based Methods}}  \\ \hline\hline
PacSum  \cite{HZheng2019PacSum}                                          & 43.2       & 14.3       & 28.5      & -       & -      & -      & -      & -      & -  & 40.3      & 17.6      & 24.9    \\  \hline
PMI  \cite{VPadmakumar2021unsupervised}                                          & 40.5       & 13.2       & 19.8      & -      & -      & -      & 37.8     & 13.4      & 29.9  & 36.7      & 14.5      & 23.3   \\  
\hline  \hline
\multicolumn{13}{|l|}{\textbf{Supervised Deep Learning based Method}}  \\ \hline\hline
MatchSum \cite{zhong2020extractive}                                          & 46.2       & 16.5       & 41.9      & -      & -      & -    & 41.2      & 14.9     & 36.8  & 44.2       & 20.6       & 40.4    \\  \hline
PEGASUS \cite{JZhang2020pegasus}                                          & 47.5       & 18.7       & 24.9      & 57.3      & 40.2      & 45.8    & 45.1      & 19.6     & 27.4  & 44.2       & 21.5       & 41.1    \\  \hline
\end{tabular}}
\caption{Comparisons between our OTExtSum and the state-of-the-art methods across different categories. The highest scores are \textbf{bold}, and the second highest ones are \underline{underlined}.
} 
\label{tab:rouges}
\end{table*}
\textit{CNN/DailyMail (CNN/DM)} \cite{KHermann2015Teaching} is the standard single-document datasets with manually-written summaries.
\textit{Multi-News} \cite{AFabbri2019Multi} is a multi-document dataset which summarises multiple news articles. We concatenate the multiple articles as a single input. \textit{BillSum} \cite{AKornilova2019billsum} is a dataset for law document summarization, which contains long state bill documents. \textit{PubMed} \cite{ACohan2018Discourse} is a scientific article dataset that uses the abstract section as the ground-truth summary and the long body section as the document. 
Table \ref{tab:dataset} shows an overview of the four datasets. The dataset details are in Appendix \ref{sec:appx_dataset}. 

While \textit{CNN/DM} contains shorter documents and summaries, the other three datasets are more challenging because they have more extended documents and summaries, thus have a higher chance to extract sentences containing redundant contents or having limited relevance to the document. 



\subsection{Implementation Details}

In terms of the pre-trained 
token embedding model, we compare the static embedding model Word2Vec and the contextual embedding models BERT and GPT2. The details of hyperparameter settings and software used are in Appendix \ref{sec:appx_hyperparameter} and \ref{sec:appx_software}. 



Our OTExtSum is compared against 
LEAD \cite{ASee2017Get}, ORACLE \cite{RNallapati2017Summarunner}, the state-of-the-art non-learning based methods and the recent unsupervised learning-based methods. LEAD and ORACLE are standard baselines in the summarisation task.
LEAD baseline extracts the first several sentences of a document as a summary. ORACLE baseline greedily extracts the sentences that maximise the ROUGE-L score based on the reference summary. We compare with the results of strong non-learning-based methods, including LSA \cite{YGong2001LSA}, TextRank \cite{RMihalcea2004Textrank}, and  LexRank \cite{GErkan2004LexRank}.
Their results on MultiNews, BillSum, PubMed, and CNN/DM are from  \cite{AFabbri2019Multi}, \cite{AKornilova2019billsum}, \cite{ACohan2018Discourse}, and \cite{VPadmakumar2021unsupervised} respectively. For an informative reference, we report recent unsupervised learning-based methods, including 
PacSum \cite{HZheng2019PacSum}, which its released model was trained on the news domain, and PMI \cite{VPadmakumar2021unsupervised}, of which the released models were trained on the news and science domains. Their results on CNN/DM are from \cite{VPadmakumar2021unsupervised}. Their results on MultiNews, BillSum, and PubMed are  evaluated on the datasets with the corresponding released models from the same domains. And we include the results of the state-of-the-art supervised learning-based methods with extractive approach MatchSum from  \cite{zhong2020extractive}, and those with abstractive approach PEGASUS from \cite{JZhang2020pegasus}.

\subsection{Quantitative Analysis} 
\label{sec:quantitative}
The commonly used ROUGE metric \cite{CLin2004Rouge} is also adopted for our quantitative analysis. It evaluates the content consistency between the generated summary and the reference summary. In detail, ROUGE-n scores measure the number of overlapping n-grams between the generated summary and the reference summary. 
A ROUGE-L score considers the longest common subsequence 
between the generated summary and the reference summary. 

\textbf{Performance Overview.} The experimental results of OTExtSum on the four datasets are listed in Table \ref{tab:rouges} 
in terms of ROUGE-1, ROUGE-2 and ROUGE-L F-scores.  We observed that the BS strategy could generally achieve better optimisation results than the BIP strategy. It is in line with our design understanding that beam search can better reach the global optimum. Whereas, the two strategies achieve similar results in CNN/DM, which could be because CNN/DM has fewer document sentences and lower budget, thus fewer possible solutions and easier to find the optimum.

OTExtSum outperforms the state-of-the-art non-learning based methods and is comparable to the learning-based methods. Note that the state-of-the-art methods usually optimise at the sentence level, whilst OTExtSum is based on the summary level OT evaluation, by which the quality of the resulting summaries is improved. 

We  observed  that  OTExtSum obtains significantly better ROUGE scores than the baseline methods on Multi-News, BillSum and PubMed, while the improvement is not that significant on CNN/DM . When the summary is more extended, such as these three more challenging datasets, the summary sentences are more likely to have redundant content. That is,  even summary-level optimisation is more difficult to achieve, our OTExtSum demonstrates higher improvements. 

OTExtSum is a non-learning based method, and training is not required. Unlike learning-based methods, it is not limited by the training data domain and can be used for different domains. Experimental results demonstrate generalisation ability of OTExtSum over news, law, and science domains.



\textbf{Effects of Token Embeddings Models.} OTExtSum is dependent on a pre-trained token embedding method. Specifically, the token embedding model affects the cost matrix \(C\) and the tokenisation, thus the frequency vector,  of the document.  We examine how different token embedding models would affect the performance of OTExtSum by comparing static embedding model Word2Vec, and contextual embedding models  BERT and GPT2. 

The results on most of the datasets indicate that a more advanced contextual embedding model such as BERT and GPT2 is more effective than a static embedding model Word2Vec. It is in line with the intuitive understanding that a more representative model with adequate training samples often approximates better token embeddings and representation. 
Despite that, the performance of OTExtSum with Word2Vec is surprisingly competitive. 

\textbf{Effects on Stop-words.} 
We investigate the impact of stop-words on the performance of OTExtSum. As shown in Table \ref{tab:ablation_rouge} in Appendix \ref{appx_abl}, the effect varies slightly across the datasets, and may not much influence the ROUGE scores. It could be because text summarisation does not generally depend on stop-words. A side benefit of removing the stop-words is reducing the vocabulary size and thus the computation time of OT.

\textbf{Effects on Distance Measurement.}
We examine how the distance measurement of the cost matrix would impact the performance of OTExtSum. As shown in Table \ref{tab:ablation_rouge} in Appendix \ref{appx_abl}, cost matrix based on the cosine distance and the Euclidean distance usually yield similar ROUGE scores.


\subsection{Interpretable Visualisation}

OTExtSum is able to provide an interpretable visualisation of the summarisation procedure. 
Figure \ref{fig:transport} in Appendix \ref{sec:appx_vis_example} illustrates the transport plan heatmap, which 
indicates the transportation of semantic contents 
between tokens in the document and its resulting summary. 
The higher the intensity, the more the semantic content of a particular document token is covered by a summary token. 

\subsection{Qualitative Analysis}

Figure \ref{fig:examplesummary_multinews} ,  \ref{fig:examplesummary_billsum} ,  \ref{fig:examplesummary_pubmed},   and \ref{fig:examplesummary_cnndm} in Appendix \ref{sec:appx_example} compare the summaries produced by OTExtSum and TextRank. 
TextRank extracted sentences that are salient on their own yet redundant when combined to form a summary. In comparison, OTExtSum is able to compose summaries that have higher semantic coverage and less redundant content.

\section{Conclusion}
\label{cha:conclusion}
In this paper, we have presented OTExtSum, the first optimal transport-based optimisation method for extractive text summarisation. It aims to identify an optimal subset of sentences for producing a summary that achieves high semantic coverage of the document by minimising the Wasserstein distance between the semantic distributions of the document and the summary. 
It helps obtain a summary from a global perspective and provides an interpretable visualisation of extraction results. 
In addition, OTExtSum does not require computationally expensive training. The comprehensive experiments demonstrate the effectiveness of OTExtSum, which is generalisable over various document domains. 
In our future work, we will explore other OT solvers for extractive summarisation.
\clearpage
\bibliography{anthology,custom}
\bibliographystyle{acl_natbib}

\clearpage
\appendix
\section{Dataset Details}
\label{sec:appx_dataset}

We followed \cite{zhong2020extractive} to set \(B\) for CNN/DM, PubMed and Multi-News, and used the average number of sentences in the summaries to set \(B\) for BillSum since this is a common
practice in the literatures \cite{SNarayan2018Ranking}. These datasets were obtained from a source, namely HuggingFace Datasets \footnote{https://huggingface.co/docs/datasets/}. 

Since OTExtSum does not require training, for a fair comparison, all experimental results are reported on the test splits of the four datasets only.

\section{Hyperparameter Details}
\label{sec:appx_hyperparameter}
For the hyperparameter settings of the BIP strategy, the number of iteration $T$ was set to 200, $\alpha$ was set to 1, and it used the SGD optimiser \cite{sutskever2013SGD} with learning rate 0.1. For the BS strategy, the beam width \(K\) was set to 5 \footnote{We chose the beam width in line with a common practice in the literature \cite{meister2020if}}.

\section{Software and Hardware Used}
\label{sec:appx_software}
We obtained the pre-trained  Word2vec (Google News 300 dimension) from GENSIM \footnote{https://radimrehurek.com/gensim/index.html}, and the contextual embedding models BERT (base version) and GPT2 from HuggingFace \footnote{https://huggingface.co}.
To compute the Wasserstein distances, we adopted GENSIM, the POT \footnote{https://pythonot.github.io} and GeomLoss \cite{feydy2019geomloss} libraries. List of stop-words was from NLTK library \footnote{https://www.nltk.org}. Our experiments were run on a GeForce GTX 1080 GPU card. We obtain our ROUGE scores by using the pyrouge package \footnote{https://pypi.org/project/pyrouge/}.

\newpage

\section{Example of Interpretable Visualisation}
\label{sec:appx_vis_example}

\begin{figure}[!h]
\includegraphics[width=0.48\textwidth]{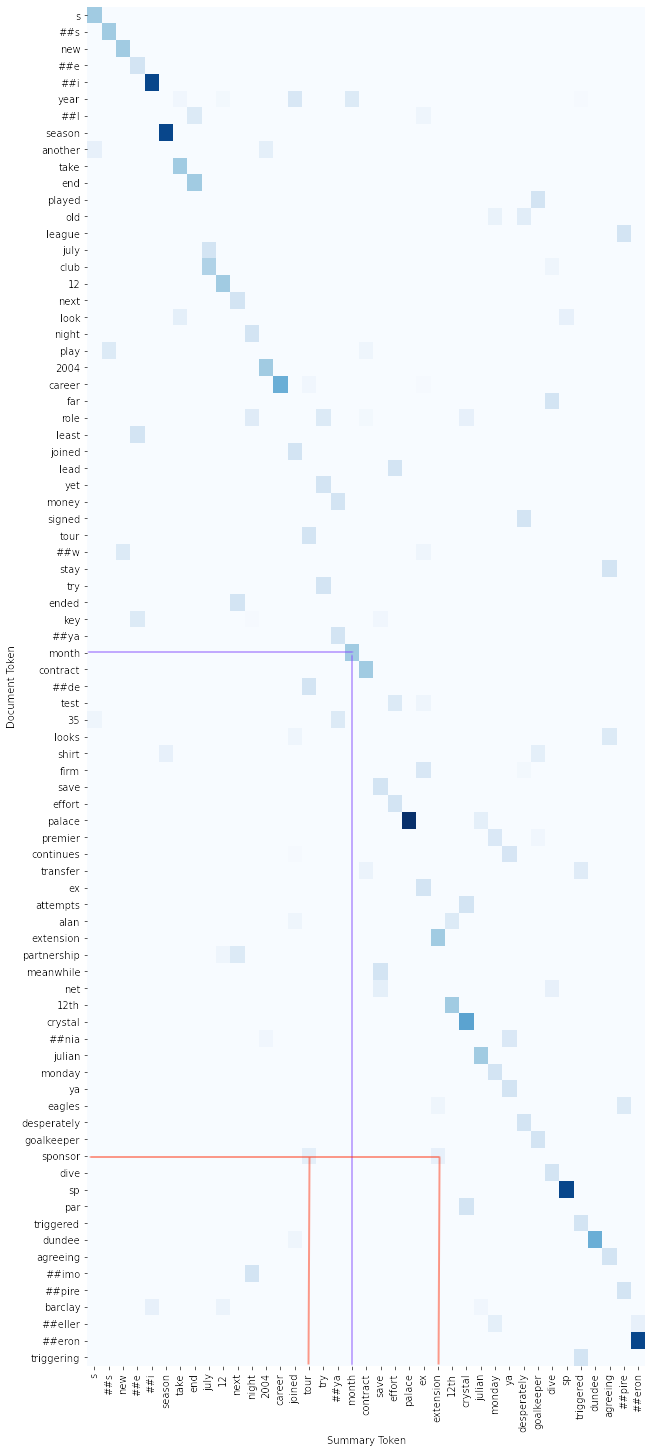}

\caption{Interpretable visualisation of the OT plan from a source document to a resulting summary on the CNN/DM dataset. The higher the intensity, the more the semantic content of a particular document token is covered by a summary token. \textcolor{violet}{Purple} line highlights the transportation from the document to the summary of semantic content of token \textcolor{violet}{``month''}, which appears in both the document and the summary. \textcolor{red}{Red} line highlights how the semantic content of token \textcolor{red}{``sponsor''}, which appears in the document only but not the summary, are transported to token \textcolor{red}{``tour''} and \textcolor{red}{``extension''}, which are semantically closer and have lower transport cost, and thus achieve a minimum transportation cost in the OT plan.  }
\label{fig:transport}
\end{figure}

\section{Ablation Studies}
\label{appx_abl}

\begin{minipage}{\textwidth}

\small
\resizebox{.98\textwidth}{!}{
\begin{tabular}{|l||ccc|ccc|ccc|ccc|}
\hline
\multicolumn{1}{|l||}{\textbf{Method}} & \multicolumn{3}{c|}{\textbf{Multi-News}} & \multicolumn{3}{c|}{\textbf{BillSum}} & \multicolumn{3}{c|}{\textbf{PubMed}} & \multicolumn{3}{c|}{\textbf{CNN/DM}}\\
\multicolumn{1}{|l||}{}                                                              & ROUGE-1     & ROUGE-2     & ROUGE-L    & ROUGE-1    & ROUGE-2    & ROUGE-L    & ROUGE-1    & ROUGE-2   & ROUGE-L   & ROUGE-1    & ROUGE-2   & ROUGE-L \\ \hline\hline
\textbf{Euc. \textbackslash wo s.w. }                                            & 43.1       & 13.9       & 22.5      & 37.5       & \textbf{19.7}      & 32.6      & 39.8      & 13.6      & 32.2  & \textbf{34.5}      & \textbf{12.8}      & \textbf{27.8}    \\  \hline
\textbf{Cos. \textbackslash wo s.w.}                                       & 43.1       & 13.9       & 22.5      & \textbf{39.0}      & 19.5      & \textbf{33.6}      & 39.8      & 13.6      & 32.3  & \textbf{34.4}      & \textbf{12.4}      & \textbf{27.7}    \\ \hline
\textbf{Euc. \textbackslash w s.w. }                                            & 43.4       & \textbf{14.4}       & \textbf{23.4}      & 36.9       & 19.6      & 32.2      & \textbf{40.6}      & \textbf{13.8}      & \textbf{33.0}  & 34.1      & 12.1      & 27.1    \\  \hline
\textbf{Cos. \textbackslash w s.w.}                                            & \textbf{43.9}       & 14.2       & 23.1      & 38.1      & 19.6      & 33.0      & \textbf{40.6}      & 13.6      & 32.9  & 34.1      & 12.1      & 27.1   \\  \hline

\end{tabular}}
\captionsetup{type=table}
\caption{Ablation studies of OTExtSum based on the BS optimisation strategy and pre-trained BERT tokeniser.  Euc. denotes the Euclidean distance and Cos. denotes the cosine distance.  s.w. denotes stop-words.  
} 
\label{tab:ablation_rouge}

\end{minipage}

\section{Generation Samples}
\label{sec:appx_example}

\begin{minipage}{0.98\textwidth}

Below are the generation samples of OTExtSum and TextRank. In general, OTExtSum based summary contains less redundant content and provides higher semantic coverage with the same number of extracted sentences.

\centering
\includegraphics[width=.98\textwidth]{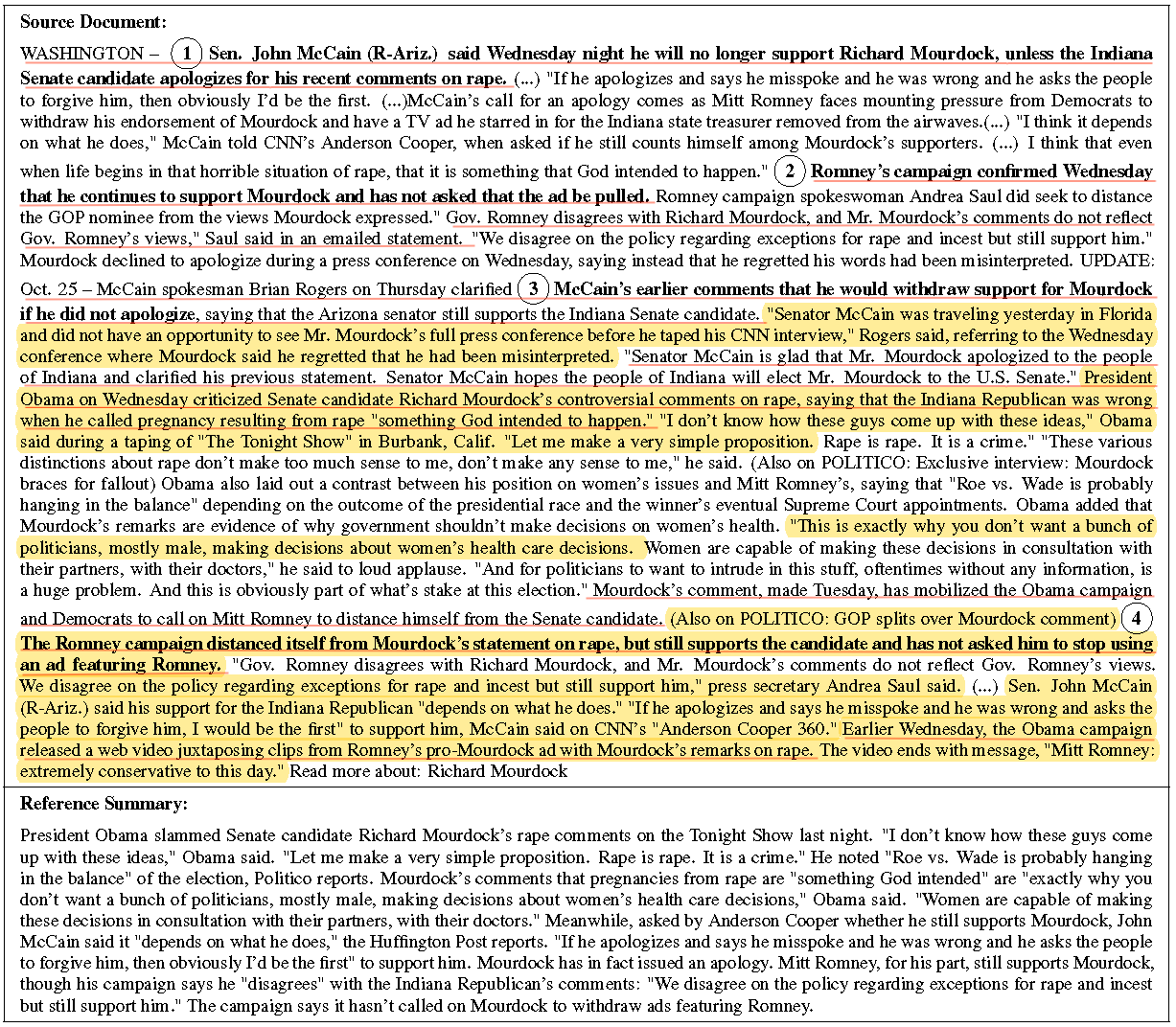}
\captionsetup{type=figure}
\caption{A sample summary comparison on the \textbf{Multi-News} dataset. OTExtSum based summary sentences are \sethlcolor{yellow} \hl{highlighted in yellow colour}. TextRank based summary sentences are \setulcolor{red} \ul{underlined in red colour}.  TextRank extracted redundant contents, specifically the part \circled{1} is duplicated with the part \circled{3}, and the part \circled{2} is duplicated with the part \circled{4}. The summary generated by OTExtSum has ROUGE-1 F-Score of 65.21 and Semantic Coverage Score of 0.93, while the summary generated by TextRank has ROUGE-1 F-Score of 44.87 and Semantic Coverage Score of 0.89.  Semantic Coverage Score of the ground-truth summary is 0.89. }
\label{fig:examplesummary_multinews}
\end{minipage}

\clearpage

\begin{minipage}{0.98\textwidth}
\centering
\includegraphics[width=.98\textwidth]{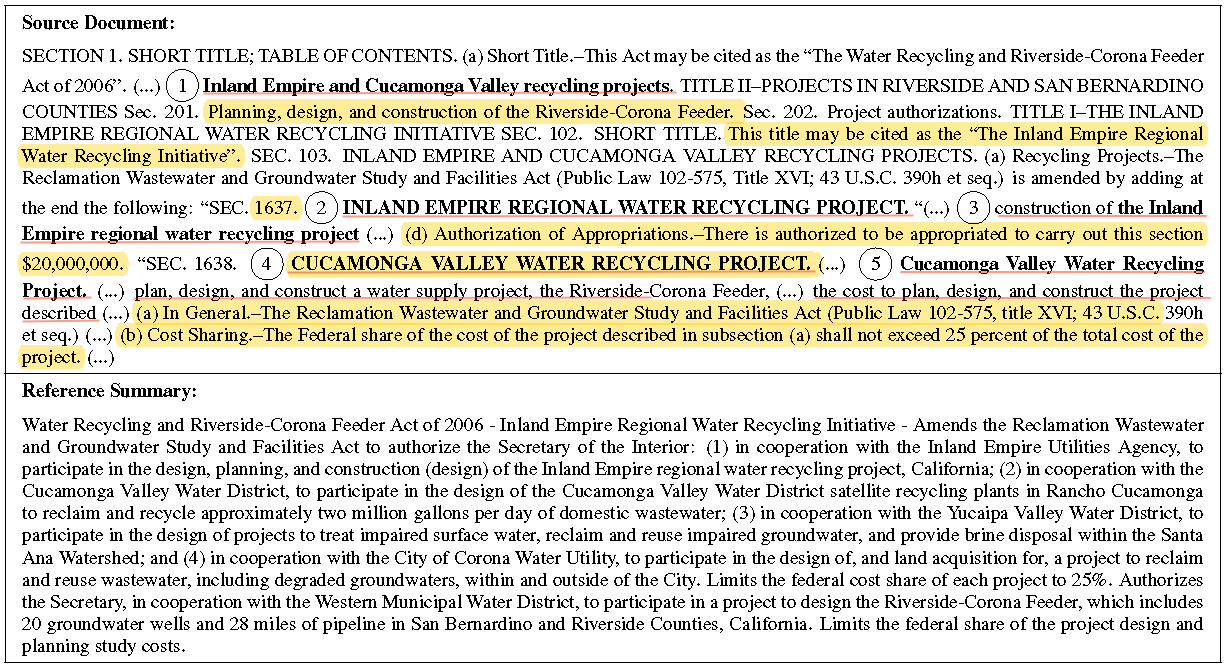}
\captionsetup{type=figure}
\caption{A sample summary comparison on the \textbf{BillSum} dataset. OTExtSum based summary sentences are \sethlcolor{yellow} \hl{highlighted in yellow colour}. TextRank based summary sentences are \setulcolor{red} \ul{underlined in red colour}.  TextRank extracted redundant contents, specifically the part \circled{1}, \circled{2}  \circled{3}, \circled{4}, and \circled{5} are duplicated. The summary generated by OTExtSum has ROUGE-1 F-Score of 44.2 and Semantic Coverage Score of 0.92, while the summary generated by TextRank has ROUGE-1 F-Score of 33.2 and Semantic Coverage Score of 0.77. Semantic Coverage Score of the ground-truth summary is 0.84.}
\label{fig:examplesummary_billsum}

\centering
\includegraphics[width=.98\textwidth]{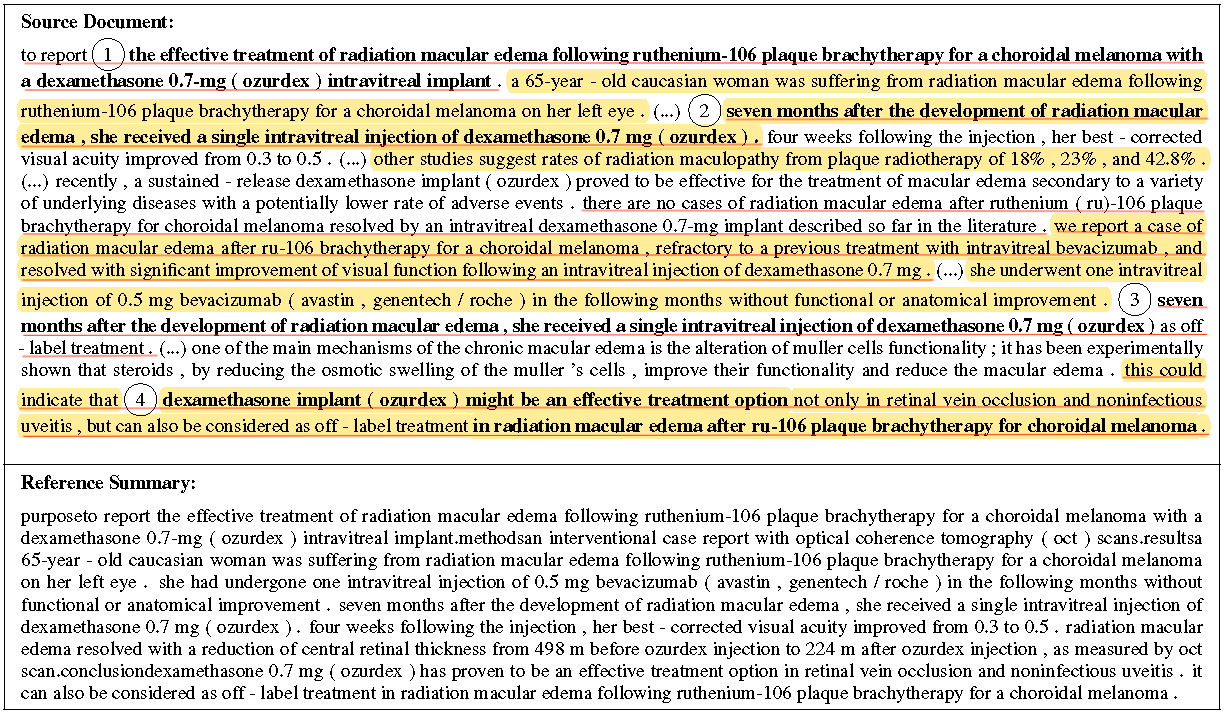}
\captionsetup{type=figure}
\caption{A sample summary comparison on the \textbf{PubMed} dataset. OTExtSum based summary sentences are \sethlcolor{yellow} \hl{highlighted in yellow colour}. TextRank based summary sentences are \setulcolor{red} \ul{underlined in red colour}.  TextRank extracted redundant contents, specifically the part \circled{1} is duplicated with the part \circled{4}, and the part \circled{2} is duplicated with the part \circled{3}. The summary generated by OTExtSum has ROUGE-1 F-Score of 73.1 and Semantic Coverage Score of 0.92, while the summary generated by TextRank has ROUGE-1 F-Score of 66.0 and Semantic Coverage Score of 0.89.  Semantic Coverage Score of the ground-truth summary is 0.91. }
\label{fig:examplesummary_pubmed}

\end{minipage}

\clearpage

\begin{minipage}{0.98\textwidth}
\centering
\includegraphics[width=.98\textwidth]{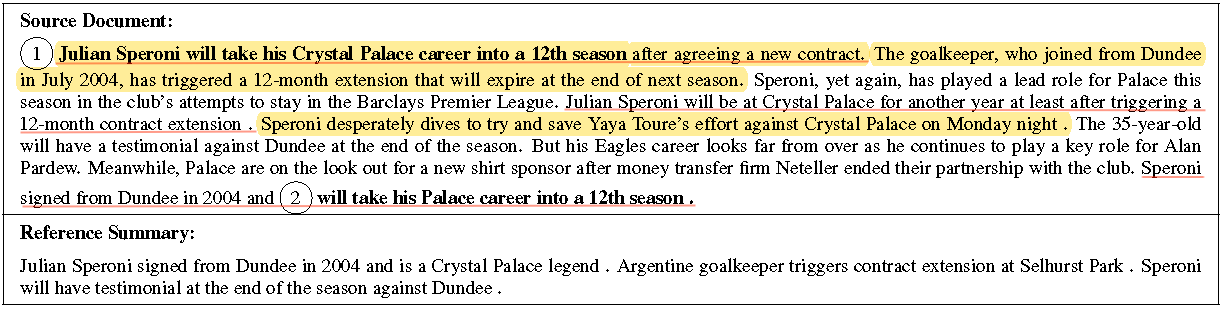}
\captionsetup{type=figure}
\caption{A sample summary comparison on the \textbf{CNN/DM} dataset. OTExtSum based summary sentences are \sethlcolor{yellow} \hl{highlighted in yellow colour}. TextRank based summary sentences are \setulcolor{red} \ul{underlined in red colour}.  TextRank extracted redundant contents, specifically the part \circled{1} is duplicated with the part \circled{2}. The summary generated by OTExtSum has ROUGE-1 F-Score of 50.5 and Semantic Coverage Score of 0.89, while the summary generated by TextRank has ROUGE-1 F-Score of 35.7 and Semantic Coverage Score of 0.83. Semantic Coverage Score of the ground-truth summary is 0.80.}
\label{fig:examplesummary_cnndm}

\end{minipage}

\end{document}